\documentclass[10pt]{article} 
\usepackage[preprint]{tmlr}

\usepackage{hyperref}
\usepackage{url}
\usepackage{graphicx}
\usepackage{booktabs}

\title{Measuring the Marginal Value of Verifier-Based RL\\Beyond Supervised Fine-Tuning}

\author{\name Amit Sagtani and Saloni Garg \email saloni8996@gmail.com \\
      \addr Santa Clara, California, USA}

\def\month{11}
\def\year{2026}
\def\openreview{\url{https://github.com/saloni-garg/gpu-train-debug}}

\begin{document}

\maketitle

\begin{abstract}
Reinforcement learning from verifiable rewards (RLVR) is usually evaluated as an alternative to supervised fine-tuning (SFT), or as a stage that follows it, with the implicit assumption that the RL stage contributes gains of its own. We measure that contribution directly, in a controlled setting where both stages can be run to completion and compared exactly. We built gpu-train-debug, a deterministic benchmark of 1,877 synthetic GPU and LLM training failure diagnosis tasks across 30 archetypes, graded by a programmatic verifier with partial credit, and calibrated so that a mid-tier model scores in a 0.60 to 0.75 band. A pilot produced a surprise: 1,200 gold demonstrations nearly solved the benchmark (SFT 0.785 on the hard split), while 15 steps of GRPO from the base model did not move it (0.314 vs 0.311 base). We therefore amended the study, in writing and before the full run, to ask the sharper question: what does verifier-based RL add beyond demonstrations? At the pre-registered scale (60 GRPO steps, learning rate 1e-6), the answer is: no detectable improvement in-distribution (+0.22 points, 95\% CI [-0.06, +0.51]), and a modest but measurable improvement under distribution shift (+0.33 points on a held-out OOD split, CI [+0.02, +0.68]). SFT accounted for most of the available improvement, while verifier-based RL produced only modest additional gains under distribution shift. The residual errors SFT leaves are mostly systematic, identical across samples, which likely reduces the within-group learning signal available to GRPO. A full 60-step cold-start GRPO arm confirms the pilot at the full horizon. The whole study cost \$8.69 of compute and is fully reproducible from archived generations, frozen configs, and tagged code.
\end{abstract}

\section{Introduction}

Post-training budgets are allocated between two broad purchases: demonstrations (human-written or teacher-generated examples used for SFT) and verifier-based RL (sampling from the policy and rewarding what a checker accepts). The standard pipeline runs SFT first and RL second \citep{lambert2024tulu3, deepseekai2025r1}, and improvements are usually reported for the pipeline as a whole, or for RL against the base model. The marginal contribution of the RL stage, over and above the demonstrations it follows, is rarely isolated, and it is the quantity that decides how the next dollar should be spent. This distinction matters because demonstrations and RL both consume post-training compute and annotation budgets, yet their individual returns are rarely measured under controlled conditions.\footnote{The experimental program reported here was designed and executed, and this manuscript drafted, with the assistance of a large language model acting under the authors' direction. All pre-registered decisions, analyses, and numerical claims were verified by the authors against the archived experimental artifacts.}

This paper isolates it in a setting designed for exactly that measurement:

\begin{itemize}
\item A task family with a deterministic verifier, so reward is not a learned proxy and cannot drift.
\item Gold demonstrations for the same distribution the RL stage trains on, so SFT and RL see the same task universe.
\item A procedurally generated OOD split, never used in training, tuning, or the pilot, so generalization claims have a clean endpoint.
\item A pre-registered analysis with claim scope and strength wording fixed in advance, so the result is reported at the size it actually is.
\end{itemize}

The study has two stages, and we report both, because the pivot between them is part of the evidence. Stage 1 (the pilot) asked whether RL can learn the task from reward alone. The answer was no at short horizon, and, unexpectedly, that demonstrations nearly saturate the benchmark. Stage 2 (the full run, under a written amendment) asked what RL adds after demonstrations. Its answer is small, precisely bounded, and investigated through a mechanistic analysis. We think this pattern, a strong SFT result followed by a marginal RL result, is common in practice and under-reported, because the intermediate comparison (SFT vs SFT+RL, with intervals) is seldom the headline.

The contribution is a package rather than a single artifact:

\begin{enumerate}
\item A carefully engineered verifier benchmark (gpu-train-debug, public, deterministic, calibrated), built for training-failure diagnosis (Section 2).
\item A methodology for designing RL-friendly verifier tasks: answer-leak hardening, a target difficulty band for the model class under study, fine-grained partial credit, and an OOD split kept pristine for confirmatory use (Sections 2 and 7).
\item A controlled empirical comparison of SFT and verifier-based RL, run as a two-stage study whose question was amended in writing after the pilot, with endpoints, intervals, and claim language pre-registered before the confirmatory run (Sections 3 to 5).
\item An unexpected near-negative result, with a headroom decomposition and a rollout-agreement analysis that bound and plausibly explain its size (Section 6).
\item A fully reproducible pipeline: archived generations, frozen and corrected configs, tagged code, and a total compute cost under \$10 (Section 9).
\end{enumerate}

\section{The environment: gpu-train-debug}

Each task presents a broken training configuration, profiler-style summaries, and log excerpts for a GPU or LLM training job. The agent must return a structured JSON answer: the fault code(s), intermediate quantities it computed (for example the true global batch size implied by the config), and corrected config fields. A deterministic grader recomputes the gold values and awards partial credit per objective. There is no learned reward model and no LLM judge anywhere in the loop.

Thirty archetypes cover single-fault diagnosis, computed quantities, interacting faults (where one fault's fix changes another's correct value), two-fault tasks, and trap tasks (where the salient log line is a red herring and the true fault is mundane). Splits are procedurally generated with fixed seeds and deduplicated: train 1,200, hard 200, test 236, OOD 241. The OOD split uses held-out surface forms and archetype variants; it was not evaluated, trained on, or tuned against at any point before the confirmatory run, a fact we verified by searching every pilot artifact for its identifier before the run.

Two design decisions matter for interpreting the results. First, the reward distribution was engineered, not merely defined: an early version leaked answers through prompt metadata and a keyword heuristic scored 1.000 on it; after hardening, the same heuristic scored 0.296, and hosted evaluation of a mid-tier model (gpt-4.1-mini) landed at 0.676, inside the 0.60 to 0.75 band we targeted. The point of the band is that a useful training environment should be neither saturated nor impossible for the model class under study. Second, partial credit is fine-grained: most tasks decompose into several independently checked objectives, so scores concentrate in (0.5, 1.0) rather than at 0 or 1. Section 6 shows this property is central to what RL could and could not do here.

The environment is publicly released (v0.4.1, MIT license) as a reinforcement learning environment for hosted evaluation and training \citep{primeintellect2025hub}; the environment, grader, and generators are available at \url{https://github.com/saloni-garg/gpu-train-debug}.

\section{Stage 1: the pilot, and the surprise}

The original question was the conventional one: on this benchmark, which learns better from the environment, SFT on gold answers or GRPO \citep{shao2024deepseekmath} on verifier reward, each from the base model (Qwen2.5-3B-Instruct \citep{qwen2025qwen25}, pinned revision)? The pilot used shortened schedules (SFT 1 epoch; GRPO 15 steps) under a frozen protocol, with the hard split as the readout.

\begin{table}[h]
\caption{Pilot results, hard split (mean reward, 3 rollouts per task)}
\label{tab:pilot}
\begin{center}
\begin{tabular}{lr}
\multicolumn{1}{c}{\bf ARM} & \multicolumn{1}{c}{\bf MEAN} \\ \hline
Base model & 0.311 \\
GRPO from base, 15 steps & 0.314 \\
SFT, 1 epoch & 0.785 \\
\end{tabular}
\end{center}
\end{table}

GRPO's training rewards showed no trend across the 15 steps (first-five mean 0.308, last-five 0.305), no divergence, and no format collapse; it simply did not learn detectably. SFT, meanwhile, gained +47.5 points, broadly across tiers (computed 0.801, interacting 0.740, two-fault 0.921), and on the test split it fixed the trap tier completely (0.100 to 1.000). One epoch of demonstrations had nearly solved a benchmark that reward alone had not dented.

Two readings were possible. The weak reading is that 15 steps is too few, which is true and worth one controlled arm. The strong reading is that the interesting question had moved: once demonstrations capture a task this efficiently, the burden of proof shifts to the RL stage to show marginal value. We adopted the strong reading, with the weak reading preserved as an exploratory arm (Section 4).

\section{Stage 2: the amended, pre-registered study}

Amendment 1 (written and adopted before any full-run training) restated the primary question: when does verifier-based RL add value beyond high-quality supervised demonstrations? It fixed the following before the run:

\begin{itemize}
\item Arms: A base (reference), B SFT (frozen 2-epoch protocol), C SFT followed by GRPO (the primary arm), D base followed by GRPO at the full 60 steps (exploratory; exists so the pilot's cold-start result cannot be dismissed as a horizon artifact).
\item Co-primary hypotheses: H1$'$ (C exceeds B on pooled interacting and computed tiers, hard split) and H1$''$ (the same on the OOD split), each tested by a paired bootstrap over tasks, 10,000 resamples, 95\% CI excluding zero.
\item A secondary concentration hypothesis (any gains sit in interacting and computed, not diagnosis or two-fault) and an equivalence hypothesis on the diagnosis tier (within plus or minus 0.03).
\item Endpoint discipline: test split descriptive only, because the pilot had observed it; OOD confirmatory, because nothing had touched it.
\item An operational SFT gate: arm C starts only if arm B reproduces (hard mean at least 0.70 and no tier far below its pilot value), so an expensive RL run never starts from a broken checkpoint.
\item Claim scope and calibration, verbatim commitments: a null means ``no detectable gain at the pre-registered scale (60 steps, lr 1e-6)'', not ``RL adds nothing''; effects under about a point with CIs excluding zero are reported as ``modest but measurable''; strong language is reserved for effects of at least 5 points.
\end{itemize}

GRPO used group size 8, 16 prompts per step, temperature 0.9, clip 0.2, KL coefficient 0.0, and a hand-rolled implementation whose log-probability computation was unit-tested against an independent recomputation. Evaluation for every arm used the frozen protocol (temperature 0.7, 3 rollouts per task, final checkpoints only) and saved every generation. The executed configuration is fully reconstructible from the archived code: frozen v1 configs are kept as provenance, corrected v2 configs govern the run, and a self-test asserts the training scripts' constants equal them. The one substantive correction this process caught: a nominal 1024-token sequence limit under which every training label would have been masked and the SFT loss NaN by construction; measured prompt lengths required 1664.

\begin{figure}[t]
\begin{center}
\includegraphics[width=\linewidth]{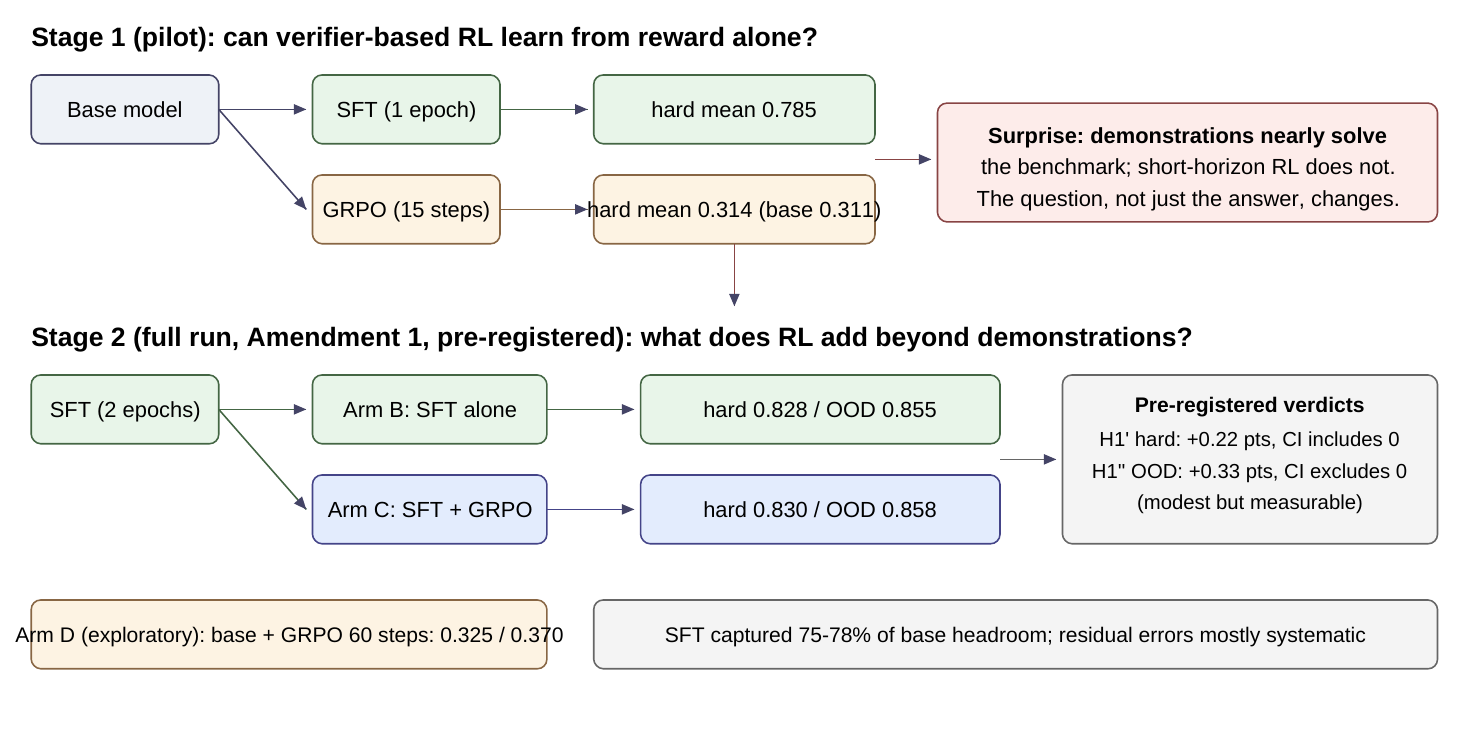}
\end{center}
\caption{The two-stage design. Stage 1 (pilot) compared SFT and short-horizon GRPO from the base model; its surprise (demonstrations nearly solve the benchmark) changed the question. Stage 2 (Amendment 1, pre-registered) measures the marginal value of GRPO after SFT, with a cold-start arm retained as an exploratory control.}
\label{fig:design}
\end{figure}

\section{Results}

The SFT gate passed decisively (hard mean 0.828 against a 0.70 threshold; the second epoch improved on the pilot's 0.785), and all arms completed without divergence or instability.

\begin{table}[h]
\caption{Mean reward by arm and split (3 rollouts per task)}
\label{tab:main}
\begin{center}
\begin{tabular}{lrrrr}
\multicolumn{1}{c}{\bf SPLIT} & \multicolumn{1}{c}{\bf BASE} & \multicolumn{1}{c}{\bf SFT (B)} & \multicolumn{1}{c}{\bf SFT+GRPO (C)} & \multicolumn{1}{c}{\bf BASE+GRPO (D)} \\ \hline
Hard (n=200) & 0.309 & 0.828 & 0.830 & 0.325 \\
OOD (n=241) & 0.340 & 0.855 & 0.858 & 0.370 \\
Test (n=236, descriptive) & 0.338 & 0.856 & 0.857 & 0.361 \\
\end{tabular}
\end{center}
\end{table}

\begin{figure}[h]
\begin{center}
\includegraphics[width=0.92\linewidth]{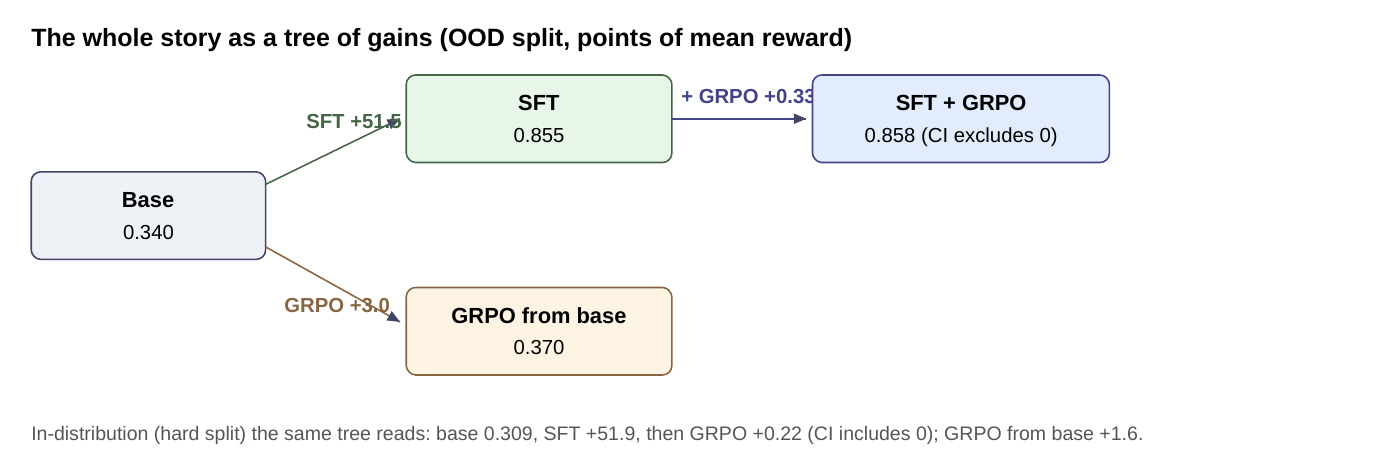}
\end{center}
\caption{The study as a tree of gains on the OOD split: demonstrations contribute +51.5 points, the RL stage that follows them +0.33, and RL without demonstrations +3.0. On the hard split the same tree reads +51.9, +0.22, and +1.6.}
\label{fig:gains}
\end{figure}

Pre-registered tests (C minus B, paired bootstrap):

\begin{itemize}
\item H1$'$ (hard, interacting+computed pooled, n=180): +0.22 points, CI [-0.06, +0.51]. Includes zero: no detectable gain in-distribution.
\item H1$''$ (OOD, pooled, n=168): +0.33 points, CI [+0.02, +0.68]. Excludes zero: per the pre-committed wording, modest but measurable.
\item Concentration: the OOD difference sits in interacting (+0.51, CI [+0.02, +1.11]) and two-fault (+0.50, CI [+0.00, +1.17]); computed is +0.14 with a CI including zero. On hard, interacting is +0.25 (CI [+0.00, +0.54]).
\item Diagnosis equivalence: the hard split contains no diagnosis-tier tasks by construction, so the test was evaluated where the tier exists. OOD diagnosis (n=24): difference exactly 0.00, CI [0.00, 0.00], equivalent within the margin.
\item Arm D (exploratory): +1.6 (hard) and +3.0 (OOD) over base at the full 60 steps, roughly 50 points below SFT. The pilot's cold-start conclusion survives the full horizon.
\end{itemize}

Per-archetype results (Appendix Tables~\ref{tab:arch-hard} and~\ref{tab:arch-ood}) show B and C within a point of each other almost everywhere, including the weakest archetype (zero\_batch: B 0.583, C 0.589 on hard; 0.583 and 0.607 on OOD) and the saturated ones (several at 1.000 for both).

\section{Analysis: headroom, flips, and why so little moved}

\textbf{Headroom decomposition.} Define headroom for a reference policy as the mean over tasks of (1 minus its per-task score): the most any successor could add. Against the base model, SFT captured 51.9 of 69.1 available points on hard (75\%) and 51.5 of 66.0 on OOD (78\%). What remained for any post-SFT method was 17.2 points on hard and 14.5 on OOD, and the realized RL differences against that remainder were the +0.22 and +0.33 points of Section 5 (about 1 to 2\% of the remainder). Cold-start GRPO, against the base model's much larger headroom, realized +1.6 and +3.0 points. The decomposition's point is accounting, not emphasis: the demonstrations took the large share of what was recoverable, and the question the rest of this section asks is why the remainder was so hard for RL to reach.

\textbf{Flip analysis.} Arms B and C differ at all on only 17 of 200 hard tasks and 24 of 241 OOD tasks; no task moved by half a point or more in either direction; the largest win is +16.7 points (one OOD zero\_batch task) and the largest regression $-$8.3 points (one hard task). On tasks SFT had saturated, C lost exactly 0.00: the RL stage did no damage. It also, nearly everywhere, did nothing.

\textbf{Rollout agreement, a mechanism.} After SFT, no task sits at the floor: every per-task mean is at least 0.5, split between saturated tasks (20\% hard, 32\% OOD) and partial-credit tasks. On the non-saturated tasks, SFT's three evaluation rollouts return identical scores on 95 of 160 (hard) and 89 of 163 (OOD): the residual errors are predominantly systematic, the model missing the same objective the same way on every sample, rather than sampling noise around a sometimes-correct answer. Group-relative RL draws its signal from within-group reward differences: a group whose samples all earn the same score contributes zero advantage by construction, and groups clustered at one fractional value contribute small ones. We therefore advance a mechanistic hypothesis, not a proven claim: the systematic character of the residual errors likely reduced the learning signal available to group-relative policy optimization in this setting, substantially reducing the variance in group-relative advantages that GRPO had to work with. The hypothesis is consistent with what arm C's training showed (batch rewards steady in a 0.77 to 0.94 band, roughly 45\% of completions perfect, and a final policy nearly identical in behavior to its initialization), and with the tier pattern of Section 5. It is measured at evaluation time, on this grader, whose partial credit makes identical fractional scores common. It would further predict that RL's marginal value grows when residual errors are stochastic rather than systematic, a condition demonstrations are unusually good at eliminating; testing that prediction directly, by constructing tasks with stochastic residuals, is future work.

\section{Discussion}

For practitioners, the actionable reading is about sequencing and measurement. If demonstrations for a task distribution are obtainable at reasonable cost, our results argue for buying them first and demanding that any subsequent RL stage justify itself against the SFT checkpoint, with intervals, on untouched data, before scaling it. The fashionable comparison, RL against base, overstates the RL stage's contribution by exactly the demonstrations' share, here 51.5 of the 51.8 points the full pipeline gains over base on OOD.

For environment builders, the headroom analysis suggests a design target beyond a calibration band: an environment intended for RL should retain headroom that is stochastic, not only systematic. A benchmark that SFT saturates in one epoch, or whose residuals are deterministic, will show small RL effects no matter how the reward is shaped; we note our own v0.4 calibration targeted the base-model band, not the post-SFT residual structure, and a v2 of the environment would add archetypes whose residual errors are sampling-sensitive.

Methodologically, the study is a small argument for pilots whose results are allowed to change the question, provided the change is written down, the new endpoints are chosen for their untouched status, and the claim language is fixed before the confirmatory run. The pivot is visible in our artifacts (the frozen original protocol, the amendment, the two tagged releases) rather than reconstructed after the fact.

\section{Limitations and threats to validity}

We list the threats explicitly, because several are the first objections a reader should raise:

\begin{itemize}
\item Model scale: a single 3B base model (Qwen2.5-3B-Instruct). Larger models may extract more from RL, or saturate differently under SFT; nothing here measures that.
\item RL scale: 60 GRPO steps at lr 1e-6 with group size 8 and one training seed, a deliberately conservative, pre-registered budget. The claim is scoped to this scale by construction; longer training is the most obvious way the result could change, and arm D bounds only the cold-start version of that objection.
\item Single RL algorithm: GRPO only. Other RLVR variants (different advantage estimators, KL control, rejection-sampling hybrids) may use the same reward stream more effectively.
\item Deterministic verifier: the grader is programmatic with dense partial credit. Learned or sparse verifiers change both the reward distribution and the residual-error structure, and the mechanism hypothesis of Section 6 is stated for this grader.
\item Single domain: GPU and LLM training-failure diagnosis, with synthetic tasks from one generator family. The OOD split varies surface forms and archetype parameters, so ``distribution shift'' here does not mean a new task domain.
\item Measurement details: 3 rollouts per task at evaluation (rollout noise is part of the estimand as frozen); diagnosis-tier equivalence testable only on OOD, since the hard split contains no diagnosis tasks; the trap tier is small (n=4 on OOD); demonstrations are gold answers from the task generator, so demonstration quality is at its ceiling, which favors SFT relative to settings where demonstrations are scarce or noisy.
\end{itemize}

Larger RL budgets, larger models, lower-quality demonstrations, and sparser rewards could each move the marginal value substantially, in either direction. The paper's claim is the measured one: at this scale, in this setting, with these controls.

\section{Reproducibility}

Everything needed to check or rerun the study is archived: the environment (v0.4.1, MIT), the code at two tagged releases (v1.0-pretraining, the frozen pre-run state, and v1.1-results), the v2 executable configs, input checksums verified on the training machine, all 8,124 evaluation generations with per-objective scores, training and reward logs, the analysis scripts, and this paper's tables generated by them. The full artifact set is archived in the public repository. Model checkpoints were intentionally not retained; they are reproducible from the archived code, configs, pinned model revision, and seeds in about 5.5 GPU-hours. Total compute cost: pilot \$3.25, full run \$5.44, on a single rented A100.

\bibliography{main}
\bibliographystyle{tmlr}

\appendix
\section{Per-archetype results}
\begin{table}[h]
\caption{Per-archetype mean reward, hard split (3 rollouts per task)}
\label{tab:arch-hard}
\begin{center}
\begin{tabular}{lrrrrr}
\multicolumn{1}{c}{\bf ARCHETYPE} & \multicolumn{1}{c}{\bf N} & \multicolumn{1}{c}{\bf BASE} & \multicolumn{1}{c}{\bf SFT} & \multicolumn{1}{c}{\bf SFT+GRPO} & \multicolumn{1}{c}{\bf BASE+GRPO} \\ \hline
accum\_and\_steps & 15 & 0.168 & 0.889 & 0.878 & 0.197 \\
accum\_steps\_lr & 15 & 0.297 & 0.848 & 0.857 & 0.297 \\
batch\_from\_components & 15 & 0.237 & 0.794 & 0.794 & 0.245 \\
checkpoint\_retention & 15 & 0.480 & 0.772 & 0.778 & 0.500 \\
ckpt\_batch\_accum & 15 & 0.111 & 0.836 & 0.836 & 0.113 \\
lr\_at\_step & 15 & 0.412 & 0.817 & 0.839 & 0.478 \\
optimizer\_batch & 15 & 0.308 & 0.778 & 0.778 & 0.336 \\
packing\_time & 15 & 0.350 & 0.889 & 0.889 & 0.372 \\
parallel\_batch & 15 & 0.222 & 0.844 & 0.844 & 0.222 \\
shard\_size & 15 & 0.639 & 1.000 & 1.000 & 0.641 \\
training\_time & 15 & 0.116 & 0.744 & 0.739 & 0.118 \\
two\_lora\_vocab & 5 & 0.547 & 1.000 & 1.000 & 0.615 \\
two\_oom\_clipping & 5 & 0.300 & 0.835 & 0.835 & 0.320 \\
two\_sampler\_accum & 5 & 0.258 & 0.895 & 0.910 & 0.297 \\
two\_v100\_dataloader & 5 & 0.533 & 1.000 & 1.000 & 0.538 \\
zero\_batch & 15 & 0.231 & 0.583 & 0.589 & 0.226 \\
\end{tabular}
\end{center}
\end{table}

\begin{table}[h]
\caption{Per-archetype mean reward, ood split (3 rollouts per task)}
\label{tab:arch-ood}
\begin{center}
\begin{tabular}{lrrrrr}
\multicolumn{1}{c}{\bf ARCHETYPE} & \multicolumn{1}{c}{\bf N} & \multicolumn{1}{c}{\bf BASE} & \multicolumn{1}{c}{\bf SFT} & \multicolumn{1}{c}{\bf SFT+GRPO} & \multicolumn{1}{c}{\bf BASE+GRPO} \\ \hline
accum\_and\_steps & 14 & 0.178 & 0.865 & 0.868 & 0.228 \\
accum\_steps\_lr & 14 & 0.280 & 0.873 & 0.871 & 0.306 \\
batch\_from\_components & 14 & 0.245 & 0.768 & 0.768 & 0.248 \\
bf16\_v100 & 2 & 1.000 & 1.000 & 1.000 & 1.000 \\
checkpoint\_retention & 14 & 0.469 & 0.756 & 0.756 & 0.500 \\
checkpointing & 2 & 0.100 & 1.000 & 1.000 & 0.100 \\
ckpt\_batch\_accum & 14 & 0.117 & 0.812 & 0.821 & 0.125 \\
clipping & 2 & 0.400 & 1.000 & 1.000 & 0.400 \\
dataloader & 2 & 0.700 & 1.000 & 1.000 & 0.742 \\
device\_mismatch & 2 & 1.000 & 1.000 & 1.000 & 1.000 \\
effective\_batch & 2 & 0.217 & 0.775 & 0.775 & 0.217 \\
fp16\_scaler & 2 & 0.100 & 1.000 & 1.000 & 0.100 \\
lora\_frozen & 2 & 1.000 & 1.000 & 1.000 & 1.000 \\
lr & 2 & 1.000 & 1.000 & 1.000 & 1.000 \\
lr\_at\_step & 14 & 0.319 & 0.839 & 0.845 & 0.430 \\
oom\_batch & 2 & 0.792 & 0.775 & 0.775 & 0.792 \\
optimizer\_batch & 14 & 0.296 & 0.768 & 0.768 & 0.337 \\
packing\_time & 14 & 0.358 & 0.917 & 0.917 & 0.387 \\
parallel\_batch & 14 & 0.198 & 0.875 & 0.875 & 0.204 \\
sampler & 2 & 0.550 & 1.000 & 1.000 & 0.700 \\
shard\_size & 14 & 0.618 & 1.000 & 1.000 & 0.640 \\
training\_time & 14 & 0.106 & 0.756 & 0.756 & 0.123 \\
trap\_nan\_clipping & 2 & 0.100 & 1.000 & 1.000 & 0.100 \\
trap\_oom\_checkpointing & 2 & 0.100 & 1.000 & 1.000 & 0.100 \\
two\_lora\_vocab & 12 & 0.508 & 1.000 & 1.000 & 0.622 \\
two\_oom\_clipping & 12 & 0.308 & 0.794 & 0.800 & 0.306 \\
two\_sampler\_accum & 12 & 0.333 & 0.913 & 0.925 & 0.355 \\
two\_v100\_dataloader & 9 & 0.562 & 1.000 & 1.000 & 0.600 \\
vocab & 2 & 0.792 & 1.000 & 1.000 & 0.833 \\
zero\_batch & 14 & 0.193 & 0.583 & 0.607 & 0.195 \\
\end{tabular}
\end{center}
\end{table}

\end{document}